\documentclass[a4paper,runningheads]{llncs}
\usepackage[latin9]{inputenc}
\PassOptionsToPackage{vlined, ruled}{algorithm2e}
\usepackage{color}
\usepackage{array}
\usepackage{float}
\usepackage{wrapfig}
\usepackage{multirow}
\usepackage{algorithm2e}
\usepackage{amsmath}
\usepackage{graphicx}
\PassOptionsToPackage{normalem}{ulem}
\usepackage{ulem}
\usepackage[unicode=true,pdfusetitle,
 bookmarks=true,bookmarksnumbered=false,bookmarksopen=false,
 breaklinks=false,pdfborder={0 0 1},backref=false,colorlinks=true]
 {hyperref}

\makeatletter

\pdfpageheight\paperheight
\pdfpagewidth\paperwidth

\providecommand{\tabularnewline}{\\}

\usepackage{graphicx}
\usepackage{tikz}
\usepackage{comment}
\usepackage{amsmath,amssymb} % define this before the line numbering.
\LinesNumbered
\usepackage{color}

\usepackage[accsupp]{axessibility}  % Improves PDF readability for those with disabilities.

\titlerunning{Black-box Few-shot Knowledge Distillation} 
\authorrunning{Dang Nguyen et al.} 

\makeatother

\begin{document}

\title{Black-box Few-shot Knowledge Distillation}

\author{Dang Nguyen, Sunil Gupta, Kien Do, Svetha Venkatesh}

\institute{Applied Artificial Intelligence Institute (A\textsuperscript{2}I\textsuperscript{2}),
Deakin University, Geelong, Australia\\
\email{$\{$d.nguyen, sunil.gupta, k.do, svetha.venkatesh$\}$@deakin.edu.au}}
\maketitle
\begin{abstract}
Knowledge distillation (KD) is an efficient approach to transfer the
knowledge from a large ``teacher'' network to a smaller ``student''
network. Traditional KD methods require lots of \textit{labeled} training
samples and a \textit{white-box} teacher (parameters are accessible)
to train a good student. However, these resources are not always available
in real-world applications. The distillation process often happens
at an external party side where we do not have access to much data,
and the teacher does not disclose its parameters due to security and
privacy concerns. To overcome these challenges, we propose a black-box
few-shot KD method to train the student with \textit{few unlabeled}
training samples and a \textit{black-box} teacher. Our main idea is
to expand the training set by generating a diverse set of out-of-distribution
synthetic images using MixUp and a conditional variational auto-encoder.
These synthetic images along with their labels obtained from the teacher
are used to train the student. We conduct extensive experiments to
show that our method significantly outperforms recent SOTA few/zero-shot
KD methods on image classification tasks. The code and models are
available at: \url{https://github.com/nphdang/FS-BBT}

\end{abstract}

\section{Introduction\label{sec:Introduction}}

Despite achieving many great successes in real-world applications
\cite{guo2019survey,sreenu2019intelligent,zhang2019deep}, deep neural
networks often have millions of weights to train, thus require heavy
computation and storage \cite{pouyanfar2018survey}. To make deep
neural networks smaller and applicable to real-time devices, especially
for edge devices with limited resources, knowledge distillation (KD)
methods have been proposed \cite{hinton2015distilling,ahn2019variational,gou2021knowledge}.

The main goal of KD is to transfer the knowledge from a large pre-trained
network (called \textit{teacher}) to a smaller network (called \textit{student})
so that the student can perform as well as the teacher \cite{hinton2015distilling,tian2020contrastive}.
Most of existing KD methods follow the idea introduced by Hinton et
al. \cite{hinton2015distilling}, which suggests to use both the ground-truth
labels and the teacher's predictions as training signals for the student.
The intuition behind this approach is that if the student network
not only learns from its training data but also is guided by a powerful
teacher network pre-trained on a large-scale data, then the student
will improve its classification accuracy.

The success of existing KD methods relies on two strong assumptions.
First, the student's training set must be \textit{very large and labeled}
(it is usually the same as the teacher's training set) \cite{ahn2019variational,hinton2015distilling,kim2018paraphrasing,tian2020contrastive}.
Second, the teacher is a \textit{white-box }model so that the student
has access to the teacher's internal details (e.g. gradient, parameters,
feature maps, logits) \cite{adriana2015fitnets,yim2017gift,chen2019data,kimura2018fewshot}.
However, these assumptions rarely hold in real-world applications.
Typically, the distillation happens at an external party side where
we can only access to few unlabeled samples. For example, DeepFace
\cite{taigman2014deepface} developed by Facebook was trained on 4
million non-public facial images. For distilling a student network
from DeepFace, an external party may not have access to the face database
used by Facebook due to various reasons including privacy. Instead,
its training set would typically comprise of a few thousands images
that are accessible at the external party side. In some cases, the
pre-trained teacher models are \textit{black-box} i.e. they are released
without disclosing their parameters, which is often the case with
cloud-deployed machine learning web-services. For example, IBM Watson
Speech-to-Text \cite{santiago2017building} only provides its APIs
to end-users to convert audio and voice to written text.

To mitigate the demand of large training data, several few-shot KD
methods were proposed for KD with few samples \cite{kimura2018fewshot,kong2020learning},
but they still require a white-box teacher. To the best of our knowledge,
there is only one method named BBKD \cite{wang2020neural} to train
the student with few samples and a black-box teacher. BBKD uses MixUp
to synthesize training images and active learning to select the most
uncertain mixup images to query the teacher model. Although BBKD shows
significant improvements over current SOTA few/zero-shot KD methods,
it exhibits two notable limitations. First, it has to synthesize \textit{a
huge pool of candidate images}. For example, given $N=1000$ original
images, it constructs $C=10^{6}$ candidate images, and selects $M=20000$
synthetic images from $C$ to train the student. Since the number
of candidate images $C$ is very large, it requires expensive computation
and consumes large memory resource. Second, it has to train the student
multiple times until a stopping criteria. Although the student network
is smaller than the teacher network, it is still a deep neural network.
Training the student multiple times must be avoided since it costs
both resources and training time. \textit{Therefore, few-shot KD with
a black-box teacher in a resource- and time-efficient manner is an
open problem}.

\textbf{Our method.} To solve the above problem, we propose a novel
\textit{unsupervised black-box few-shot }KD method i.e. training the
student with only \textit{few unlabeled} images and a \textit{black-box}
teacher. Our method offers a resource- and time-efficient KD process,
which addresses the bottlenecks of BBKD. First, it does not need to
create any pool of candidate images; instead it directly generates
$M$ synthetic images from $N$ original images to train the student.
Second, it only trains the student network in one-pass; no active
learning is required and no multiple student models are repeatedly
created.

\begin{figure}
\begin{centering}
\includegraphics[scale=0.27]{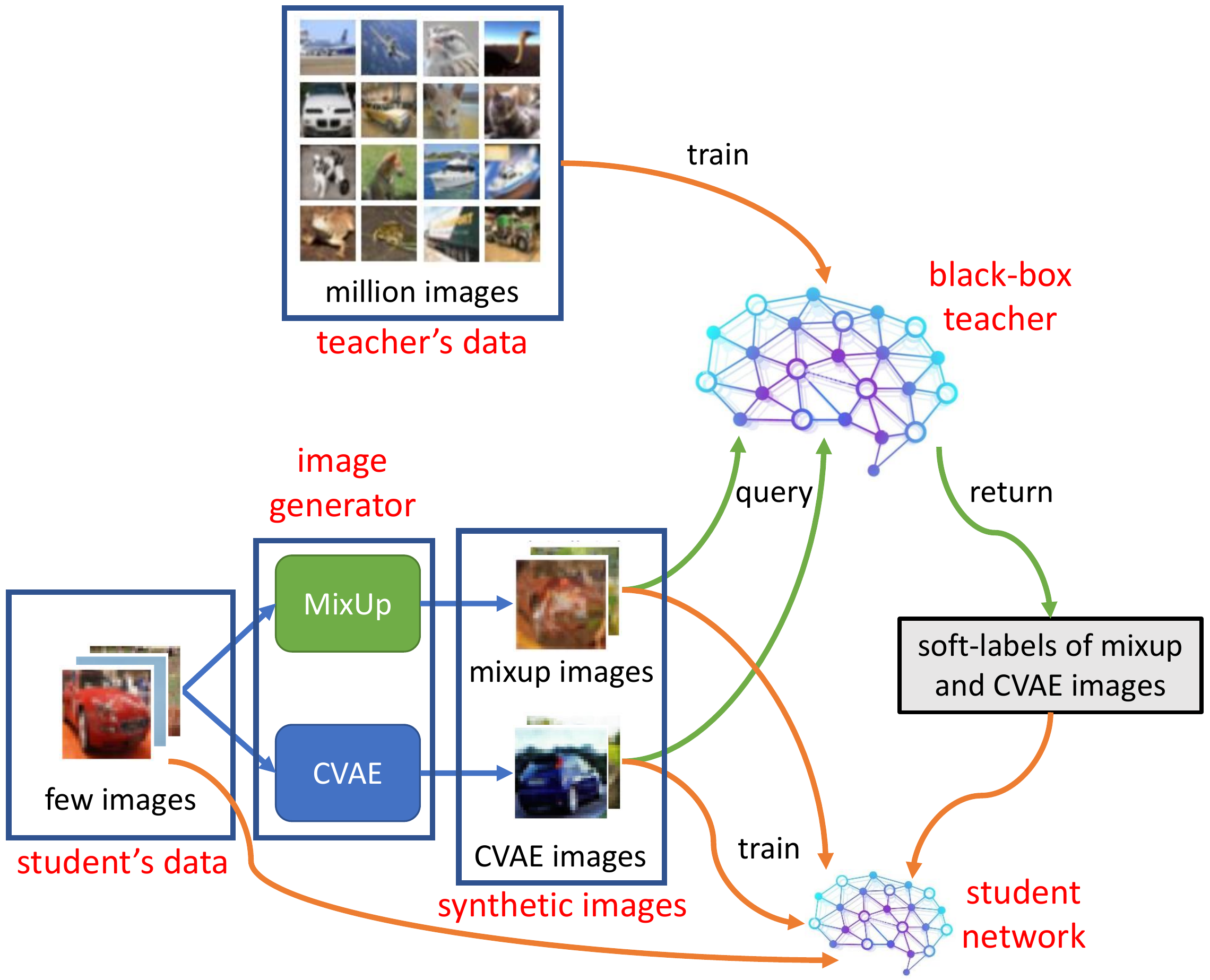}
\par\end{centering}
\caption{\label{fig:KD-FS-BBT}Knowledge distillation with few samples and
black-box teacher. Given a black-box teacher and a small set of original
images, we propose to employ MixUp method and CVAE generative model
to generate synthetic images to train the student network.}
\end{figure}

Our method has three main steps. First, we generate synthetic images
from a given \textit{small} set of original images. Second, the synthetic
images are sent to the teacher model to query their \textit{soft-labels}
(i.e. class probabilities). Finally, the original and synthetic images
along with their soft-labels are used to train the student network.
Our method is illustrated in Figure \ref{fig:KD-FS-BBT}.

The key component in our method is the image generator, where we propose
two approaches to generate synthetic images. First, we use the MixUp
method \cite{zhang2018mixup,guo2019mixup,berthelot2019mixmatch} to
synthesize a virtual image by a weighted combination of two original
images. Mixup images help us to cover the manifold of natural images.
Second, we use Conditional Variational Autoencoder (CVAE) \cite{sohn2015learning}
-- a generative model to generate additional synthetic images. While
MixUp is useful to some extent, mixup images too close to original
images do not add much value to the training data. Such disqualified
mixup images are replaced by images generated from CVAE. Using CVAE,
we can generate interpolated images i.e. the output image semantically
mixes characteristics from the original images. As a result, we can
enrich the training set and improve the diversity of training images,
which is very useful when training the student network.

\textbf{Our contribution.} To summarize, we make the following contributions.
\begin{enumerate}
\item We propose \textbf{FS-BBT} (\textit{knowledge distillation with }\textit{\uline{F}}\textit{ew
}\textit{\uline{S}}\textit{amples and }\textit{\uline{B}}\textit{lack-}\textit{\uline{B}}\textit{ox
}\textit{\uline{T}}\textit{eacher}), a novel method offers a successful
KD process even with few unlabeled training samples and a black-box
teacher model.
\item We develop an efficient approach to train the student network in resource-
and time-efficient manner, where we do not need to create a large
pool of candidate images and only train the student network one time.
\item We empirically validate our proposed method on several image classification
tasks, comparing it with both standard and SOTA few/zero-shot KD methods.
The experimental results show that our method significantly outperforms
competing baselines.
\end{enumerate}

\section{Related Works\label{sec:Related-Works}}

\textbf{Knowledge distillation.} Knowledge distillation (KD) has become
popular since Hinton et al. introduced its concept in their teacher-student
framework \cite{hinton2015distilling}. The main goal of KD is to
train a compact student network by mimicking the softmax output of
a high-capacity teacher network. Many KD methods have been proposed,
and they can be categorized into three groups: \textit{relation-based},
\textit{feature-based}, and \textit{response-based} methods. Relation-based
methods not only use the teacher's output but also explore the relationships
between different layers of teacher when training the student network.
Examples include \cite{yim2017gift,lee2019graph,passalis2020heterogeneous}.
Feature-based methods leverage both the teacher's output at the last
layer and the intermediate layers when training student network \cite{adriana2015fitnets,kim2018paraphrasing,passalis2020heterogeneous}.
Response-based methods directly mimic the final prediction of the
teacher network \cite{hinton2015distilling,chen2017learning,meng2019conditional,nguyen2021knowledge}.

\textbf{Knowledge distillation with limited data.} To successfully
train the student network, most KD methods assume that both the student's
training data and the teacher's training data are identical. For example,
\cite{hinton2015distilling,chawla2021data} pointed out that the student
only achieved its best accuracy when it had accessed to the teacher's
training data. Similarly, \cite{nayak2021effectiveness} mentioned
the typical setting in KD methods was the student network trained
on the teacher's training data. Recent SOTA methods \cite{kim2018paraphrasing,ahn2019variational,tian2020contrastive}
also trained both teacher and student networks on the same dataset.
In practice, the teacher's training data could be unavailable due
to transmission limitation or privacy while we could only collect
few samples for the student's training data. Several few/zero-shot
KD methods were developed to deal with this situation \cite{kimura2018fewshot,nayak2019zero,chen2019data,yin2020dreaming,kong2020learning}.
However, all of these methods require a \textit{white-box} teacher
to access to its internal details (e.g. gradient information, weights,
feature maps, logits...) to generate synthetic training samples. As
far as we know, only BBKD \cite{wang2020neural} requires few training
samples and zero knowledge of the teacher (i.e. \textit{black-box}
teacher). However, it is computation and resource intensive as it
requires a large pool of candidate images and extensive iterative
training.

\section{Framework\label{sec:Framework}}

\subsection{Problem definition\label{subsec:Problem-definition}}

Given a small set of \textit{unlabeled} images ${\cal X}=\{x_{i}\}_{i=1}^{N}$
and a black-box teacher $T$, our goal is to train a student $S$
on ${\cal X}$ s.t. $S$'s performance is comparable to $T$'s.

A direct solution for the above problem is to apply the standard KD
method \cite{hinton2015distilling}. We first query the teacher to
obtain the \textit{hard-label} (i.e. one-hot encoding) $y_{i}$ for
each sample $x_{i}\in{\cal X}$, and then create a labeled training
set ${\cal D}=\{x_{i},y_{i}\}_{i=1}^{N}$. Finally, we train the student
network with the standard KD loss function:
\begin{equation}
{\cal L}=\sum_{(x_{i},y_{i})\in{\cal D}}(1-\omega){\cal L}_{CE}(y_{x_{i}}^{S},y_{i})+\omega{\cal L}_{KL}(y_{x_{i}}^{S},y_{x_{i}}^{T}),\label{eq:KD-loss}
\end{equation}
where $y_{x_{i}}^{S}$, $y_{x_{i}}^{T}$, $y_{i}$ are the student's
softmax output, the teacher's softmax output, and the hard-label of
a sample $x_{i}$, ${\cal L}_{CE}$ is the cross-entropy loss, ${\cal L}_{KL}$
is the Kullback--Leibler divergence loss, and $\omega$ is a trade-off
factor to balance the two loss terms. Equation (\ref{eq:KD-loss})
does not use the \textit{temperature} factor as in Hinton\textquoteright s
KD method \cite{hinton2015distilling} since this requires access
to the pre-softmax activations (logits) of teacher, which violates
our assumption of \textquotedblleft black-box\textquotedblright{}
teacher.

Although training the student network via Equation (\ref{eq:KD-loss})
is a possible way, it is not a good solution as ${\cal X}$ only contains
very few samples while standard KD methods typically require lots
of training samples \cite{hinton2015distilling,kim2018paraphrasing,ahn2019variational,tian2020contrastive}.

\subsection{Proposed method FS-BBT\label{subsec:Proposed-method-FS-BBT}}

We propose a novel method to solve the above problem, which has three
main steps: (1) we generate mixup images from original images contained
in ${\cal X}$, (2) we replace disqualified mixup images by images
generated from CVAE, and (3) we train the student with a combination
of original, mixup, and CVAE images.

\textbf{Generating mixup images.} Our idea is to use MixUp \cite{zhang2018mixup}
-- one of recently proposed data augmentation techniques to expand
the training set ${\cal X}$.

Inspired by BBKD \cite{wang2020neural}, we generate $M$ mixup images
from $N$ original images (typically, $N\ll M$). Given two original
images $x_{i},x_{j}\in{\cal X}$, we use MixUp to generate a synthetic
image by a weighted combination between $x_{i}$ and $x_{j}$:
\begin{equation}
x_{mu}(\lambda)=\lambda x_{i}+(1-\lambda)x_{j},\label{eq:mixup-image}
\end{equation}
where the coefficient $\lambda\in[0,1]$ is sampled from a Beta distribution.

Let $X=[x_{1},x_{2},...,x_{N}]$ be the vector of original images.
We first sample two $M$-length vectors $X^{1}=[x_{1}^{1},x_{2}^{1},...,x_{M}^{1}]$
and $X^{2}=[x_{1}^{2},x_{2}^{2},...,x_{M}^{2}]$, where $x_{i}^{1},x_{i}^{2}\sim X$.
We then sample a vector $\lambda=[\lambda_{1},\lambda_{2},...,\lambda_{M}]$
from a Beta distribution, and mixup each pair of two images in $X^{1}$
and $X^{2}$ using Equation (\ref{eq:mixup-image}):
\begin{equation}
X_{mu}=\left[\begin{array}{c}
\lambda_{1}x_{1}^{1}+(1-\lambda_{1})x_{1}^{2}\\
\lambda_{2}x_{2}^{1}+(1-\lambda_{2})x_{2}^{2}\\
...\\
\lambda_{M}x_{M}^{1}+(1-\lambda_{M})x_{M}^{2}
\end{array}\right]\label{eq:X-mixup}
\end{equation}

The goal of mixing up original images is to expand the initial set
of training images ${\cal X}$ as much as possible to cover the manifold
of natural images.

\begin{wrapfigure}{o}{0.5\columnwidth}%
\begin{centering}
\includegraphics[scale=0.5]{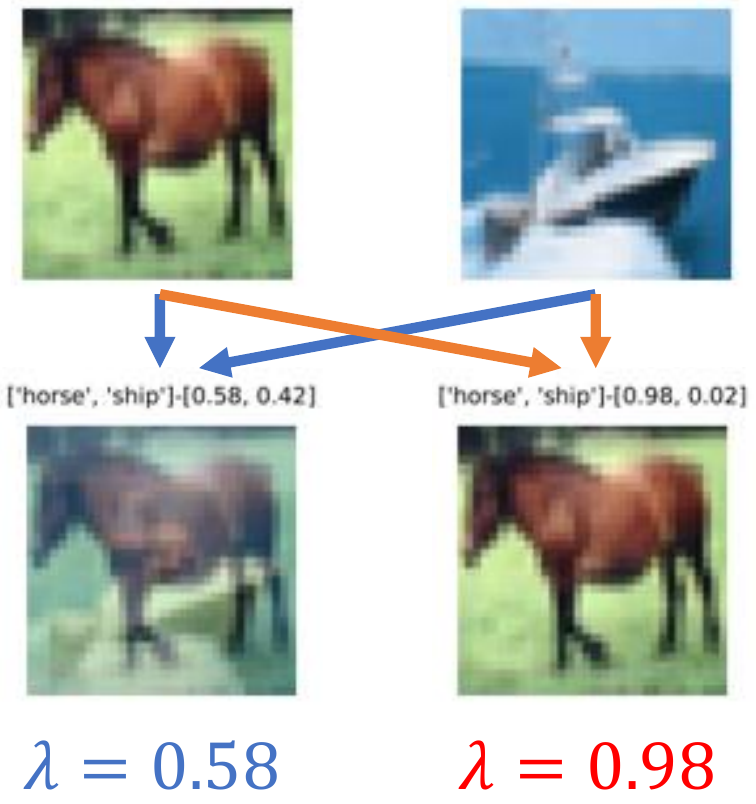}
\par\end{centering}
\caption{\label{fig:Examples-of-mixup_images}Desirable vs. disqualified mixup
images. At $\lambda=0.58$, the mixup image shows a good combination
between two original images ``horse'' and ``ship'' but at $\lambda=0.98$,
it looks almost the same as ``horse''.}
\end{wrapfigure}%

However, when mixing up two original images, there is a case that
the mixup image is very similar to one of two original images, making
it useless. This problem happens when $\lambda_{i}\approx0$ or $\lambda_{i}\approx1$.
Figure \ref{fig:Examples-of-mixup_images} shows two examples of desirable
vs. disqualified mixup images.

To remove disqualified mixup images, we set a threshold $\alpha\in[0,0.5]$,
and discard mixup images generated with coefficient $\lambda_{i}\leq\alpha$
or $\lambda_{i}\geq(1-\alpha)$.

Let $M_{1}$ be the number of remaining mixup images after we filter
out the disqualified ones. Our next step is to generate $M_{2}=M-M_{1}$
synthetic images from CVAE (we call them CVAE images).

\textbf{Generating CVAE images.} We first query the teacher model
to obtain the hard-label $y_{i}$ for each sample $x_{i}\in{\cal X}$
to create a labeled training set ${\cal D}=\{x_{i},y_{i}\}_{i=1}^{N}$.
We then train a Conditional Variational Autoencoder (CVAE) model \cite{sohn2015learning}
using ${\cal D}$ to learn the distribution of the latent variable
$z\in\mathbb{R}^{d}$, where $d$ is the dimension of $z$. CVAE is
a generative model consisting of an encoder and a decoder. We use
the encoder network to map an image along with its label $(x,y)\in{\cal D}$
to a latent vector $z$ that follows $P(z\mid y)$. From the latent
vector $z$ conditioned on the label $y$, we use the decoder network
to reconstruct the input image $x$. Following \cite{sohn2015learning},
we train CVAE by maximizing the variational lower bound objective:
\begin{align}
\log P(x & \mid y)\geq\mathbb{E}(\log P(x\mid z,y))-\text{KL}(Q(z\mid x,y),P(z\mid y)),\label{eq:cvae-loss}
\end{align}
where $Q(z\mid x,y)$ is parameterized by the encoder network that
maps input image $x$ and its label $y$ to the latent vector $z$,
$P(x\mid z,y)$ is parameterized by the decoder network that reconstructs
input image $x$ from the latent vector $z$ and label $y$, $\mathbb{E}(\log P(x\mid z,y))$
is the expected likelihood, which is implemented by a cross-entropy
loss between the input image and the reconstructed image, and $P(z\mid y)\equiv{\cal N}(0,I)$
is the prior distribution of $z$ conditioned on $y$.

After the CVAE model is trained, we can generate images via $G(z,y)$,
where $z\sim{\cal N}(0,I)$, $y$ is a label, and $G$ is the trained
decoder network.

\textbf{Covering both in-distribution and out-of-distribution samples.}
To generate $M_{2}$ CVAE images, we sample ($\frac{M_{2}}{2}$)-length
vector $z^{{\cal N}}$ from the normal distribution ${\cal N}(0,I)$
and ($\frac{M_{2}}{2}$)-length vector $z^{{\cal U}}$ from the uniform
distribution ${\cal U}([-3,3]^{d})$ (we choose the range $[-3,3]$
following \cite{higgins2017beta,gyawali2019semi}). We create vector
$z=z^{{\cal N}}\oplus z^{{\cal U}}$, where $\oplus$ is the concatenation
operator. We manually define a $M_{2}$-length vector $y_{cvae}$,
which contains the classes of generated images such that the number
of generated images for each class is equivalent. Finally, we generate
CVAE images $x_{cvae}=G(z,y_{cave})$.

The intuition behind our generation process is that: (1) Generating
images from $z^{{\cal N}}\sim{\cal N}(0,I)$ will provide \textit{synthetic
images within the distribution} of ${\cal X}$. These images are interpolated
versions of original images. (2) Generating images from $z^{{\cal U}}\sim{\cal U}([-3,3]^{d})$
will provide \textit{synthetic images out-of the distribution} of
${\cal X}$. These images are far way from the original ones, but
they are expected to better cover \textit{unseen images,} which improves
the student's generalization.

\textbf{Discussion.} One can sample $\lambda_{i}\in[\alpha,1-\alpha]$
to generate $M_{1}$ qualified mixup images, then generate $M_{2}$
CVAE images. This way requires two hyper-parameters $M_{1}$ and $M_{2}$.
While this is definitely possible, for simplicity we choose to aggregate
these two hyper-parameters into a single hyper-parameter $M$ that
controls the total number of synthetic images. In experiments, we
set the same values for $M$ as those in other few-shot KD methods
\cite{kimura2018fewshot,wang2020neural} while $M_{1}$ and $M_{2}$
are automatically computed based on $M$ and $\alpha$.

\textbf{Training the student network.} After the above steps, we obtain
two types of synthetic images -- mixup and CVAE images. We send them
to the teacher model to obtain their softmax outputs (i.e. their class
probabilities) as the \textit{soft-labels} for the images. We train
the student network with the original and synthetic images along with
their soft-labels using the following loss:
\begin{equation}
{\cal L}=\sum_{x_{i}\in{\cal X}\cup{\cal X}_{mu}\cup{\cal X}_{cvae}}{\cal L}_{CE}(y_{x_{i}}^{S},y_{x_{i}}^{T}),\label{eq:our-loss}
\end{equation}
where $y_{x_{i}}^{S}$, $y_{x_{i}}^{T}$ are the student's and the
teacher's softmax output, ${\cal X}$, ${\cal X}_{mu}$, ${\cal X}_{cvae}$
are the set of original, mixup, and CVAE images, and ${\cal L}_{CE}$
is the cross-entropy loss. Although we train the student by matching
the teacher's softmax outputs, our loss function is still applicable
in case the teacher only returns top-1 labels \cite{wang2021zero}.
Algorithm \ref{alg:The-proposed-KDFS} summarizes our proposed method
\textbf{FS-BBT}.

\begin{algorithm}
\caption{\label{alg:The-proposed-KDFS}The proposed \textbf{FS-BBT} algorithm.}

\LinesNumbered

\KwIn{$T$: pre-trained \textit{black-box} teacher network}

\KwIn{$\mathcal{X}=\{x_{i}\}_{i=1}^{N}$: \textit{unlabeled} training
set}

\KwIn{$M$: number of synthetic images}

\KwIn{$\alpha$: threshold to select mixup images}

\KwOut{$S$: student network}

\Begin{

query teacher $T$ to obtain hard-label $y_{i}$ for each $x_{i}\in{\cal X}$\;

train CVAE model using ${\cal D}=\{x_{i},y_{i}\}_{i=1}^{N}$\;

sample $\lambda=[\lambda_{1},...,\lambda_{M}]$ from a Beta distribution\;

select $M_{1}$ instances of $\lambda_{i}$ s.t. $\alpha<\lambda_{i}<1-\alpha$\;

generate $M_{1}$ mixup images ${\cal X}_{mu}$ using Eq. (\ref{eq:X-mixup})\;

compute $M_{2}=M-M_{1}$\;

sample ($\frac{M_{2}}{2}$)-length vector $z^{{\cal N}}\sim{\cal N}(0,I)$\;

sample ($\frac{M_{2}}{2}$)-length vector $z^{{\cal U}}\sim{\cal U}([-3,3]^{d})$\;

create vector $z=z^{{\cal N}}\oplus z^{{\cal U}}$\;

design $M_{2}$-length vector $y_{cvae}$ with class balance\;

generate $M_{2}$ CVAE images ${\cal X}_{cvae}=G(z,y_{cvae})$\;

query teacher $T$ to obtain soft-labels for ${\cal X}$, ${\cal X}_{mu}$,
${\cal X}_{cvae}$\;

train student $S$ with ${\cal X},{\cal X}_{mu},{\cal X}_{cvae}$
and their soft-labels using Eq. (\ref{eq:our-loss})\;

}
\end{algorithm}

\section{Experiments and Discussions\label{sec:Experiments}}

We conduct extensive experiments on five benchmark image datasets
to evaluate the classification performance of our method, comparing
it with SOTA baselines. Our main goal is to show that with the same
number of original and synthetic images, our method is much better
than existing few/zero-shot KD methods.

\subsection{Datasets}

We use five image datasets, namely MNIST, Fashion-MNIST, CIFAR-10,
CIFAR-100, and Tiny-ImageNet. These datasets were often used to evaluate
the classification performance of KD methods \cite{hinton2015distilling,kimura2018fewshot,chen2019data,nayak2019zero,wang2020neural}.

\subsection{Baselines}

We compare our method \textbf{FS-BBT} with the following baselines:
\begin{itemize}
\item \textit{Student-Alone}: the student network is trained on the student's
training data ${\cal D}$ from scratch.
\item \textit{Standard-KD}: the student network is trained with the standard
KD loss in Equation (\ref{eq:KD-loss}). We choose the trade-off factor
$\omega=0.9$, which is a common value used in KD methods \cite{hinton2015distilling,tian2020contrastive,yuan2020revisiting,ma2021undistillable}.
\item \textit{FSKD} \cite{kimura2018fewshot}: this is a few-shot KD method,
which generates synthetic training images using adversarial technique.
It requires a \textit{white-box} teacher model to generate adversarial
samples to train the student network.
\item \textit{WaGe} \cite{kong2020learning}: this few-shot KD method integrates
a Wasserstein-based loss with the standard KD loss to improve the
student's generalization.
\item \textit{BBKD} \cite{wang2020neural}: this method uses few original
images and a \textit{black-box} teacher model to train the student
model. Its main idea is to use MixUp and active learning to generate
synthetic images. Since this is the closest work to ours, we consider
BBKD as our main competitor.
\end{itemize}
To have a fair comparison, we use the same teacher-student network
architecture, the same number of original and synthetic images $N$
and $M$ as in FSKD and BBKD. We also set the same hyper-parameters
(e.g. batch size and the number of epochs) for Student-Alone, Standard-KD,
and our \textbf{FS-BBT}. We use threshold $\alpha=0.05$ to select
qualified mixup images across all experiments. In an ablation study
in Section \ref{subsec:Hyper-parameter-analysis}, we will investigate
how different values for $\alpha$ affect our method's performance.
We repeat each experiment five times with random seeds, and report
the averaged accuracy. For the baselines FSKD, WaGe, and BBKD, we
obtain their accuracy from the papers \cite{kong2020learning,wang2020neural}\footnote{This is possible because we use benchmark datasets, and the training
and test splits are fixed.}. We also compare with several well-known zero-shot KD methods in
Section \ref{subsec:Comparison-with-zero-shot}.

\subsection{Results on MNIST and Fashion-MNIST}

\subsubsection{Experiment settings.}

Following \cite{kimura2018fewshot,nayak2019zero}, we use the LeNet5
architecture \cite{lecun2015lenet} for the teacher and LeNet5-Half
(a modified version with half number of channels per layer) for the
student. We train the teacher network with a batch size of 64 and
20 epochs. As shown in Table \ref{tab:Classification-results-MNIST},
our teacher model achieves comparable accuracy with that reported
by BBKD in \cite{wang2020neural} (99.18\% vs. 99.29\% for MNIST and
90.15\% vs. 90.80\% for Fashion-MNIST). We train the student network
with a batch size of 64 and 50 epochs. We train the CVAE with feed-forward
neural networks for both encoder and decoder, using a latent dimension
of 2, a batch size of 256, and 100 (200) epochs for MNIST (Fashion-MNIST).
Following FSKD \cite{kimura2018fewshot} and BBKD \cite{wang2020neural},
we set $N=2000$ and $M=24000$ for MNIST and $N=2000$ and $M=48000$
for Fashion-MNIST.

The MNIST and Fashion-MNIST datasets have 60K training images and
10K testing images from 10 classes ($[0,1,...,9]$).

\subsubsection{Quantitative results.}

From Table \ref{tab:Classification-results-MNIST}, we can see that
our method \textbf{FS-BBT} outperforms Student-Alone and Standard-KD
on both MNIST and Fashion-MNIST. \textbf{FS-BBT} achieves 98.42\%
(MNIST) and 84.73\% (Fashion-MNIST), which is much better than Student-Alone
achieving 95.97\% and 81.37\%. With a support from the teacher model,
Standard-KD is always better than Student-Alone, for example, 83.87\%
vs. 81.37\% on Fashion-MNIST.

\begin{table}
\caption{\label{tab:Classification-results-MNIST}Classification results on
MNIST and Fashion-MNIST. ``Teacher'' indicates the accuracy of the
teacher network on the test set. ``Model'' indicates whether the
teacher network is a \textit{black-box} model. ``$N$'' shows the
number of original images used by each method. ``Accuracy'' is the
accuracy of the student network on the test set. The results of\textbf{
}FSKD, WaGe, and BBKD\protect\textsuperscript{$\star$} are obtained
from \cite{kong2020learning,wang2020neural}. ``$\star$'' means
the BBKD\protect\textsuperscript{$\star$} and \textbf{FS-BBT}\protect\textsuperscript{$\star$}
methods use the same architecture (LeNet5) for both teacher and student
networks.}

\centering{}%
\begin{tabular}{|l|l|r|c|r|r|}
\hline 
\textbf{Dataset} & \textbf{Method} & \textbf{Teacher} & \textbf{Model} & \textbf{$N$} & \textbf{Accuracy}\tabularnewline
\hline 
\hline 
\multirow{6}{*}{MNIST} & Student-Alone & - & - & 2,000 & 95.97\%\tabularnewline
\cline{2-6} 
 & Standard-KD & 99.18\% & Black & 2,000 & 95.99\%\tabularnewline
\cline{2-6} 
 & FSKD \cite{kimura2018fewshot} & 99.29\% & White & 2,000 & 80.43\%\tabularnewline
\cline{2-6} 
 & BBKD\textsuperscript{$\star$} \cite{wang2020neural} & 99.29\% & Black & 2,000 & 98.74\%\tabularnewline
\cline{2-6} 
 & \textbf{FS-BBT }(Ours) & 99.18\% & Black & 2,000 & 98.42\%\tabularnewline
\cline{2-6} 
 & \textbf{FS-BBT}\textsuperscript{$\star$} (Ours) & 99.18\% & Black & 2,000 & \textbf{98.91\%}\tabularnewline
\hline 
\multicolumn{6}{c}{}\tabularnewline
\hline 
\multirow{7}{*}{Fashion-MNIST} & Student-Alone & - & - & 2,000 & 81.37\%\tabularnewline
\cline{2-6} 
 & Standard-KD & 90.15\% & Black & 2,000 & 83.87\%\tabularnewline
\cline{2-6} 
 & FSKD \cite{kimura2018fewshot} & 90.80\% & White & 2,000 & 68.64\%\tabularnewline
\cline{2-6} 
 & WaGe \cite{kong2020learning} & 92.00\% & White & 1,000 & 85.18\%\tabularnewline
\cline{2-6} 
 & BBKD\textsuperscript{$\star$} \cite{wang2020neural} & 90.80\% & Black & 2,000 & 80.90\%\tabularnewline
\cline{2-6} 
 & \textbf{FS-BBT }(Ours) & 90.15\% & Black & 2,000 & 84.73\%\tabularnewline
\cline{2-6} 
 & \textbf{FS-BBT}\textsuperscript{$\star$} (Ours) & 90.15\% & Black & 2,000 & \textbf{86.53\%}\tabularnewline
\hline 
\end{tabular}
\end{table}

Compared with FSKD and WaGe, \textbf{FS-BBT} significantly outperforms
FSKD on both MNIST and Fashion-MNIST while \textbf{FS-BBT} is similar
with WaGe on Fashion-MNIST. 

Compared with BBKD, \textbf{FS-BBT} achieves a comparable accuracy
with BBKD on MNIST while \textbf{FS-BBT} outperforms BBKD by a large
margin on Fashion-MNIST, where our accuracy improvement is around
4\%. Since BBKD uses the same architecture LeNet5 for both teacher
and student networks, we also report the accuracy of our method with
this setting, indicated by \textbf{FS-BBT}\textsuperscript{$\star$}.
With LeNet5 for the student network, we further achieve 2\% gain (i.e.
an improvement of 6\% over BBKD) on Fashion-MNIST.

\subsection{Results on CIFAR-10 and CIFAR-100}

\subsubsection{Experiment settings.}

Following \cite{kimura2018fewshot,nayak2019zero}, we use AlexNet
\cite{krizhevsky2012imagenet} and AlexNet-Half (50\% filters are
removed) for teacher and student networks on CIFAR-10. We train the
teacher network with a batch size of 512 and 50 epochs. Our teacher
model achieves a comparable accuracy with that reported by BBKD in
\cite{wang2020neural} (84.07\% vs. 83.07\%). We train the student
network with a batch size of 128 and 100 epochs. We use ResNet-32
\cite{he2016deep} for the teacher and ResNet-20 for the student on
CIFAR-100. We train student and teacher networks with a batch size
of 16/32 and 200 epochs. For both CIFAR-10 and CIFAR-100, we train
the CVAE model with convolutional neural networks for both encoder
and decoder, using a latent dimension of 2, a batch size of 64, and
600 epochs. Like BBKD \cite{wang2020neural} and WaGe \cite{kong2020learning},
we set $N=2000$ for CIFAR-10, $N=5000$ for CIFAR-100, and $M=40000$
for both datasets.

CIFAR-10 is set of RGB images with 10 classes, 50K training images,
and 10K testing images while CIFAR-100 is with 100 classes, and each
class contains 500 training images and 100 testing images. Since neither
the accuracy reference nor the source code is available for BBKD on
CIFAR-100, we implement BBKD by ourselves, and use the same teacher
as in our method for a fair comparison.

\begin{table}
\caption{\label{tab:Classification-results-CIFAR10}Classification results
on CIFAR-10 and CIFAR-100. ``$N$'' shows the number of original
images used by each method. The results of\textbf{ }FSKD, WaGe, and
BBKD\protect\textsuperscript{$\star$} are obtained from \cite{kong2020learning,wang2020neural}.
``$\star$'' means the BBKD\protect\textsuperscript{$\star$} and
\textbf{FS-BBT}\protect\textsuperscript{$\star$} methods use the
same architecture (AlexNet) for both teacher and student networks.
``$\dagger$'' means the result is based on our own implementation.}

\centering{}%
\begin{tabular}{|l|l|r|c|r|r|}
\hline 
\textbf{Dataset} & \textbf{Method} & \textbf{Teacher} & \textbf{Model} & \textbf{$N$} & \textbf{Accuracy}\tabularnewline
\hline 
\hline 
\multirow{7}{*}{CIFAR-10} & Student-Alone & - & - & 2,000 & 54.59\%\tabularnewline
\cline{2-6} 
 & Standard-KD & 84.07\% & Black & 2,000 & 58.96\%\tabularnewline
\cline{2-6} 
 & FSKD \cite{kimura2018fewshot} & 83.07\% & White & 2,000 & 40.58\%\tabularnewline
\cline{2-6} 
 & WaGe \cite{kong2020learning} & 89.00\% & White & 5,000 & 73.08\%\tabularnewline
\cline{2-6} 
 & BBKD\textsuperscript{$\star$} \cite{wang2020neural} & 83.07\% & Black & 2,000 & 74.60\%\tabularnewline
\cline{2-6} 
 & \textbf{FS-BBT} (Ours) & 84.07\% & Black & 2,000 & 74.10\%\tabularnewline
\cline{2-6} 
 & \textbf{FS-BBT}\textsuperscript{$\star$} (Ours) & 84.07\% & Black & 2,000 & \textbf{76.17\%}\tabularnewline
\hline 
\multicolumn{6}{c}{}\tabularnewline
\hline 
\multirow{5}{*}{CIFAR-100} & Student-Alone & - & - & 5,000 & 32.85\%\tabularnewline
\cline{2-6} 
 & Standard-KD & 69.08\% & Black & 5,000 & 36.79\%\tabularnewline
\cline{2-6} 
 & WaGe \cite{kong2020learning} & 47.00\% & White & 5,000 & 20.32\%\tabularnewline
\cline{2-6} 
 & BBKD\textsuperscript{$\dagger$} \cite{wang2020neural} & 69.08\% & Black & 5,000 & 53.41\%\tabularnewline
\cline{2-6} 
 & \textbf{FS-BBT} (Ours) & 69.08\% & Black & 5,000 & \textbf{56.28\%}\tabularnewline
\hline 
\end{tabular}
\end{table}

\subsubsection{Quantitative results.}

From Table \ref{tab:Classification-results-CIFAR10} we observe the
similar results as in MNIST and Fashion-MNIST. Student-Alone does
not have a good accuracy. Standard-KD improves 4\% of accuracy over
Student-Alone with the knowledge transferred from the teacher.

On CIFAR-10, WaGe and BBKD greatly outperform FSKD, and our \textbf{FS-BBT}
is comparable with WaGe and BBKD. When we use the same architecture
AlexNet for both teacher and student as in BBKD, our variant \textbf{FS-BBT}\textsuperscript{$\star$}
is the best method, where it outperforms BBKD (the second best method)
by around 2\%. \textbf{FS-BBT}\textsuperscript{$\star$} outperforms
WaGe by around 3\% even though WaGe uses much more original training
samples than ours (5K vs. 2K), and more powerful teacher (89\% vs.
84\%).

On CIFAR-100, Student-Alone achieves low accuracy at around 32\%.
Standard-KD is better than Student-Alone around 4\% thanks to the
knowledge transferred from the teacher. Interestingly, WaGe works
very poorly (only 20.32\% of accuracy), becoming the worst method.
Its unsatisfactory performance can be a consequence of distilling
from a low-accuracy teacher. BBKD is significantly better than other
methods with an improvement around 20-30\%. Using the same number
of original and synthetic images, our method \textbf{FS-BBT} achieves
3\% gains over BBKD thanks to the CVAE images generated in Section
\ref{subsec:Proposed-method-FS-BBT}.

The above results suggest that replacing disqualified mixup images
by synthetic images generated from CVAE is an effective solution to
improve the robustness and generalization of the student network on
the unseen testing samples, as we discussed in Section \ref{subsec:Proposed-method-FS-BBT}.

\subsection{Results on Tiny-ImageNet}

\subsubsection{Experiment settings.}

We use ResNet-32 and ResNet-20 for the teacher and student. We train
teacher and student networks with a batch size of 32 and 100 epochs.
Our teacher model achieves a similar accuracy with literature \cite{bhat2021distill}
(52.02\% vs. 48.26\%). We train CVAE in the same way as in CIFAR-100.
We set $N=10000$ and $M=50000$. Tiny-ImageNet has 100K training
images, 10K testing images, and 200 classes.

\begin{table}
\caption{\label{tab:Classification-results-TinyImageNet}Classification results
on Tiny-ImageNet. ``$N$'' shows the number of original images used
by each method. ``$\dagger$'' means the result is based on our
own implementation.}

\centering{}%
\begin{tabular}{|l|l|r|c|r|r|}
\hline 
\textbf{Dataset} & \textbf{Method} & \textbf{Teacher} & \textbf{Model} & \textbf{$N$} & \textbf{Accuracy}\tabularnewline
\hline 
\hline 
\multirow{5}{*}{Tiny-ImageNet} & Student-Alone (full) & - & - & 100,000 & 48.81\%\tabularnewline
\cline{2-6} 
 & Student-Alone & - & - & 10,000 & 23.19\%\tabularnewline
\cline{2-6} 
 & Standard-KD & 52.02\% & Black & 10,000 & 35.81\%\tabularnewline
\cline{2-6} 
 & BBKD\textsuperscript{$\dagger$} \cite{wang2020neural} & 52.02\% & Black & 10,000 & 40.01\%\tabularnewline
\cline{2-6} 
 & \textbf{FS-BBT} (Ours) & 52.02\% & Black & 10,000 & \textbf{43.29\%}\tabularnewline
\hline 
\end{tabular}
\end{table}

\subsubsection{Quantitative results.}

Table \ref{tab:Classification-results-TinyImageNet} shows that Student-Alone
reaches a very low accuracy due to a large number of classes presented
in this dataset. Standard-KD is significantly better than Student-Alone
with an improvement more than 12\%. Our method \textbf{FS-BBT} achieves
3\% gains over BBKD (the second-best baseline).

We also train Student-Alone with full 100K original images and their
soft-labels provided by the teacher. This can be considered as an
upper bound of all few-shot KD methods as it uses the full set of
training images. \textbf{FS-BBT} drops only 5\% accuracy from Student-Alone
with full training data although it requires only 10\% of training
data. This proves the efficacy of our proposed framework.

\subsection{Comparison with zero-shot (or data-free) KD methods\label{subsec:Comparison-with-zero-shot}}

We also compare with several popular zero-shot KD methods, including
\textit{Meta-KD} \cite{lopes2017data}, \textit{ZSKD} \cite{nayak2019zero},
\textit{DAFL} \cite{chen2019data}, \textit{DFKD} \cite{wang2021data},
and \textit{ZSDB3KD} \cite{wang2021zero}. 

Table \ref{tab:Classification-data-free} reports the classification
accuracy on MNIST, Fashion-MNIST, and CIFAR-10. Our method is much
better than other methods on Fashion-MNIST and CIFAR-10 while it is
comparable on MNIST.

\begin{table}
\caption{\label{tab:Classification-data-free}Classification comparison with
zero-shot KD methods. The results of baselines are obtained from \cite{wang2021zero}.}

\centering{}%
\begin{tabular}{|l|c|r|r|r|}
\hline 
\textbf{Method} & \textbf{Model} & \textbf{MNIST} & \textbf{Fashion-MNIST} & \textbf{CIFAR-10}\tabularnewline
\hline 
\hline 
Meta-KD \cite{lopes2017data} & White & 92.47\% & - & -\tabularnewline
\hline 
ZSKD \cite{nayak2019zero} & White & 98.77\% & 79.62\% & 69.56\%\tabularnewline
\hline 
DAFL \cite{chen2019data} & White & 98.20\% & - & 66.38\%\tabularnewline
\hline 
DFKD \cite{wang2021data} & White & \textbf{99.08\%} & - & 73.91\%\tabularnewline
\hline 
ZSDB3KD \cite{wang2021zero} & Black & 96.54\% & 72.31\% & 59.46\%\tabularnewline
\hline 
\textbf{FS-BBT }(Ours) & Black & 98.91\% & \textbf{86.53\%} & \textbf{76.17\%}\tabularnewline
\hline 
\end{tabular}
\end{table}

\subsection{Ablation study}

As there are several components and a hyper-parameter $\alpha$ in
our method, we further conduct some ablation experiments to analyze
how each of them affects to our overall classification accuracy. We
select CIFAR-10 for this analysis.

\subsubsection{Different types of synthetic images.\label{subsec:Different-types-of}}

As described in Section \ref{subsec:Proposed-method-FS-BBT}, we generate
three types of synthetic images to train the student network. First,
we generate \textit{mixup images}. Second, we sample $z^{{\cal N}}\sim{\cal N}(0,I)$
to generate CVAE images within the distribution of the original images
(we call them \textit{CVAE-WD images}). Finally, we sample $z^{{\cal U}}\sim{\cal U}([-3,3]^{d})$
to generate CVAE images out-of the distribution of the original images
(we call them \textit{CVAE-OOD images}).

\begin{figure}[H]
\begin{centering}
\includegraphics[scale=0.3]{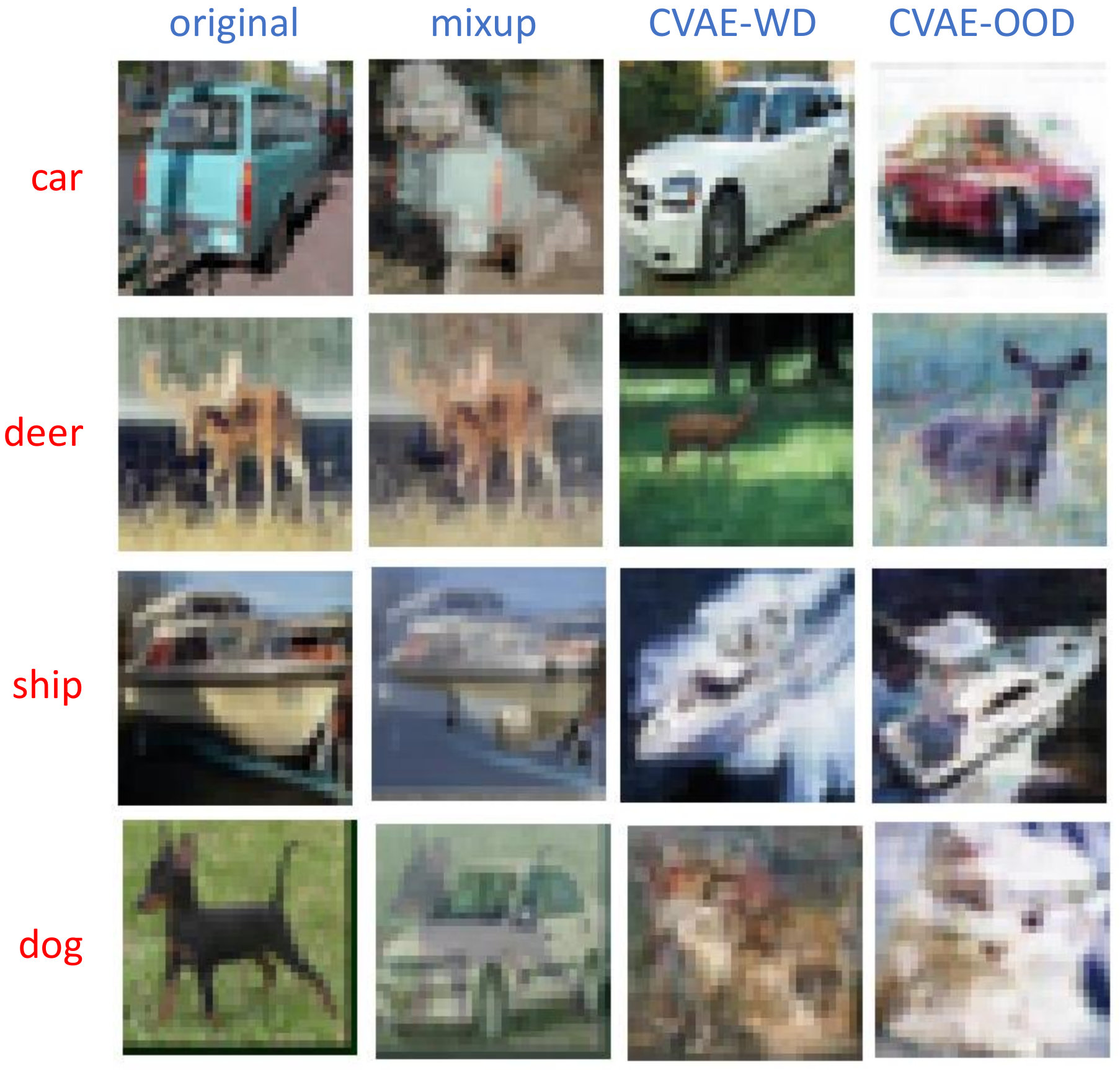}
\par\end{centering}
\caption{\label{fig:Original-synthetic-images}Original images (1\protect\textsuperscript{st}
column) and three types of synthetic images: mixup images (2\protect\textsuperscript{nd}
column), CVAE-WD images (3\protect\textsuperscript{rd} column), and
CVAE-OOD images (4\protect\textsuperscript{th} column). The text
on the left indicates the true labels of original images.}
\end{figure}

Figure \ref{fig:Original-synthetic-images} shows original images
and three types of synthetic images for four true classes ``car'',
``deer'', ``ship'', and ``dog''. Our synthetic images have good
quality, where the objects are clearly recognized and visualized.
These synthetic images provide a comprehensive coverage of real images
in the test set, resulting in the great improvement of the student
network trained on them.

Table \ref{tab:Effectiveness-of-different-synthetic} reports the
accuracy of various types of our synthetic images. The standard KD
method achieves only 58.96\% of accuracy. By utilizing mixup images,
our method achieves up to 71.67\% of accuracy. However, using solely
mixup images has disadvantages as we discussed in Section \ref{subsec:Proposed-method-FS-BBT}.
By combining mixup images with CVAE-WD images or CVAE-OOD images,
our method further improves its accuracy up to 72.60\% and 73.25\%
of accuracy respectively. Finally, when combining all three types
of synthetic images, our method achieves the best performance at 74.10\%
of accuracy.

\begin{table}
\caption{\label{tab:Effectiveness-of-different-synthetic}Effectiveness of
different types of synthetic images on our method \textbf{FS-BBT}.}

\centering{}%
\begin{tabular}{|l|c|c|c|c|c|c|c|c|}
\hline 
 & KD & \multicolumn{7}{c|}{\textbf{FS-BBT} (Ours)}\tabularnewline
\hline 
\hline 
mixup images &  & $\checkmark$ &  &  & $\checkmark$ & $\checkmark$ &  & $\checkmark$\tabularnewline
\hline 
CVAE-WD images &  &  & $\checkmark$ &  & $\checkmark$ &  & $\checkmark$ & $\checkmark$\tabularnewline
\hline 
CVAE-OOD images &  &  &  & $\checkmark$ &  & $\checkmark$ & $\checkmark$ & $\checkmark$\tabularnewline
\hline 
\textbf{Accuracy} & 58.96\% & 71.67\% & 70.26\% & 69.42\% & 72.60\% & 73.25\% & 70.63\% & \textbf{74.10\%}\tabularnewline
\hline 
\end{tabular}
\end{table}

The ablation experiments suggest that each type of synthetic images
in our method is meaningful, where it greatly improves the student's
classification performance compared to the standard KD method. By
leveraging all three types of synthetic images, our method improves
the generalization and diversity of the training set, which is very
effective for the training of the student network.

\subsubsection{Hyper-parameter analysis.\label{subsec:Hyper-parameter-analysis}}

Our method \textbf{FS-BBT} has one hyper-parameter, that is, the threshold
$\alpha$ to determine disqualified mixup images and replace them
by CVAE images (see Section \ref{subsec:Proposed-method-FS-BBT}).
We examine how the different choices of $\alpha$ affect our classification.

\begin{wrapfigure}{o}{0.5\columnwidth}%
\begin{centering}
\includegraphics[scale=0.3]{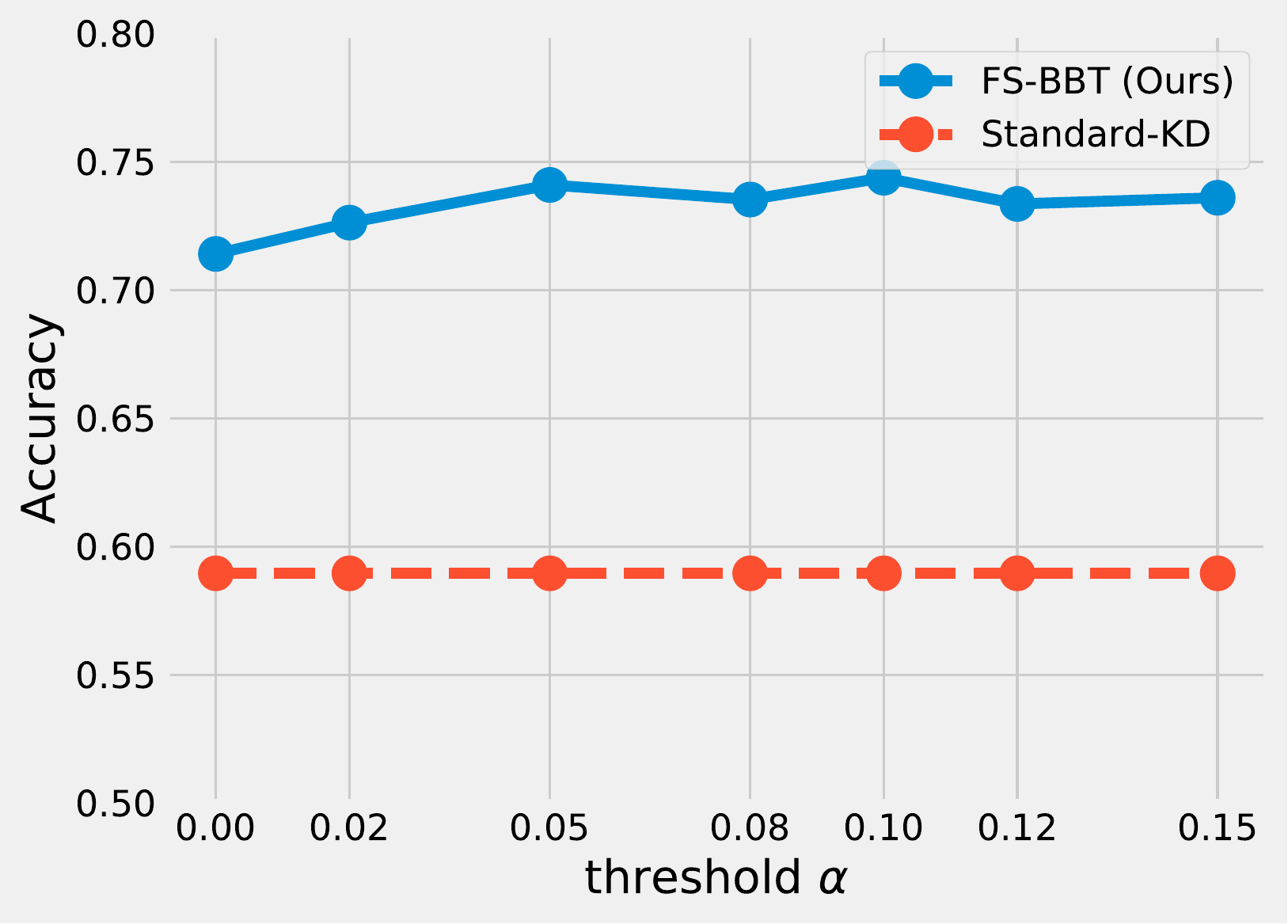}
\par\end{centering}
\caption{\label{fig:Hyper-parameter-sensitivity}\textbf{FS-BBT}'s accuracy
vs. threshold $\alpha$ on CIFAR-10.}
\end{wrapfigure}%

As shown in Figure \ref{fig:Hyper-parameter-sensitivity}, \textbf{FS-BBT}
is always better than the standard KD method regardless of $\alpha$
values. More importantly, it is stable with $\alpha\in[0.05,0.10]$,
where its accuracy just slightly changes. When $\alpha$ is too small
(i.e. $\alpha<0.05$), most of mixup images will be considered qualified
although many of them are very similar to the original images, leading
to few extra meaningful training samples added. The performance of
\textbf{FS-BBT} is decreased as expected. When $\alpha$ is too large
(i.e. $\alpha>0.10$), \textbf{FS-BBT} also slightly reduces its accuracy.
This is because many mixup images may become cluttered and semantically
meaningless due to a large proportion of two original images blended
together, making them difficult for the teacher network to label.

\section{Conclusion}

Existing standard and few/zero-shot KD methods require lots of original
training data or a white-box teacher, which are not realistic in some
cases. We present \textbf{FS-BBT} -- a novel KD method, which is
effective even with few training samples and a black-box teacher.
\textbf{FS-BBT} uses MixUp and CVAE to generate synthetic images to
train the student network. Although neither of them is new, combining
them is a novel solution to address the problem of black-box KD with
few samples. As \textbf{FS-BBT} is \textit{unsupervised}, which does
not require any ground-truth labels, it can be directly applied to
domains where labeled images are difficult to obtain e.g. medical
images. We demonstrate the benefits of \textbf{FS-BBT} on five benchmark
image datasets, where it significantly outperforms SOTA baselines.
Our work can cheaply create a white-box proxy of a black-box model,
which allows algorithmic assurance \cite{gopakumar2018algorithmic,ha2021high}
to verify its behavior along various aspects e.g. robustness, fairness,
safety, etc.

\textbf{Acknowledgment:} This research was fully supported by the
Australian Government through the Australian Research Council\textquoteright s
Discovery Projects funding scheme (project DP210102798). The views
expressed herein are those of the authors and are not necessarily
those of the Australian Government or Australian Research Council.

\bibliographystyle{splncs04}
\bibliography{reference}

\begin{thebibliography}{10}
\providecommand{\url}[1]{\texttt{#1}}
\providecommand{\urlprefix}{URL }
\providecommand{\doi}[1]{https://doi.org/#1}

\bibitem{adriana2015fitnets}
Adriana, R., Nicolas, B., Ebrahimi, S., Antoine, C., Carlo, G., Yoshua, B.:
  {Fitnets: Hints for thin deep nets}. In: ICLR (2015)

\bibitem{ahn2019variational}
Ahn, S., Hu, X., Damianou, A., Lawrence, N., Dai, Z.: Variational information
  distillation for knowledge transfer. In: CVPR. pp. 9163--9171 (2019)

\bibitem{kimura2018fewshot}
Akisato, K., Zoubin, G., Koh, T., Tomoharu, I., Naonori, U.: Few-shot learning
  of neural networks from scratch by pseudo example optimization. In: British
  Machine Vision Conference (BMVC). p.~105 (2018)

\bibitem{berthelot2019mixmatch}
Berthelot, D., Carlini, N., Goodfellow, I., Papernot, N., Oliver, A., Raffel,
  C.: {MixMatch: A Holistic Approach to Semi-Supervised Learning}. In: NIPS.
  vol.~32 (2019)

\bibitem{bhat2021distill}
Bhat, P., Arani, E., Zonooz, B.: Distill on the go: Online knowledge
  distillation in self-supervised learning. In: CVPR. pp. 2678--2687 (2021)

\bibitem{chawla2021data}
Chawla, A., Yin, H., Molchanov, P., Alvarez, J.: {Data-Free Knowledge
  Distillation for Object Detection}. In: CVPR. pp. 3289--3298 (2021)

\bibitem{chen2017learning}
Chen, G., Choi, W., Yu, X., Han, T., Chandraker, M.: Learning efficient object
  detection models with knowledge distillation. In: NIPS. pp. 742--751 (2017)

\bibitem{chen2019data}
Chen, H., Wang, Y., Xu, C., Yang, Z., Liu, C., Shi, B., Xu, C., Xu, C., Tian,
  Q.: Data-free learning of student networks. In: ICCV. pp. 3514--3522 (2019)

\bibitem{gopakumar2018algorithmic}
Gopakumar, S., Gupta, S., Rana, S., Nguyen, V., Venkatesh, S.: Algorithmic
  assurance: An active approach to algorithmic testing using bayesian
  optimisation. NIPS  \textbf{31} (2018)

\bibitem{gou2021knowledge}
Gou, J., Yu, B., Maybank, S., Tao, D.: Knowledge distillation: A survey.
  International Journal of Computer Vision  \textbf{129}(6),  1789--1819 (2021)

\bibitem{guo2019survey}
Guo, G., Zhang, N.: A survey on deep learning based face recognition. Computer
  Vision and Image Understanding  \textbf{189},  102805 (2019)

\bibitem{guo2019mixup}
Guo, H., Mao, Y., Zhang, R.: Mixup as locally linear out-of-manifold
  regularization. In: AAAI. vol.~33, pp. 3714--3722 (2019)

\bibitem{gyawali2019semi}
Gyawali, K.: {Semi-Supervised Learning by Disentangling and Self-Ensembling
  Over Stochastic Latent Space}. arXiv preprint arXiv:1907.09607  (2019)

\bibitem{ha2021high}
Ha, H., Gupta, S., Rana, S., Venkatesh, S.: {High Dimensional Level Set
  Estimation with Bayesian Neural Network}. In: AAAI. vol.~35, pp. 12095--12103
  (2021)

\bibitem{he2016deep}
He, K., Zhang, X., Ren, S., Sun, J.: Deep residual learning for image
  recognition. In: CVPR. pp. 770--778 (2016)

\bibitem{higgins2017beta}
Higgins, I., et~al.: beta-vae: Learning basic visual concepts with a
  constrained variational framework. In: ICLR (2017)

\bibitem{hinton2015distilling}
Hinton, G., Vinyals, O., Dean, J.: Distilling the knowledge in a neural
  network. arXiv preprint arXiv:1503.02531  (2015)

\bibitem{zhang2018mixup}
Hongyi, Z., Moustapha, C., Yann, D., David, L.P.: mixup: Beyond empirical risk
  minimization. In: ICLR (2018)

\bibitem{kim2018paraphrasing}
Kim, J., Park, S., Kwak, N.: Paraphrasing complex network: Network compression
  via factor transfer. In: NIPS. pp. 2760--2769 (2018)

\bibitem{kong2020learning}
Kong, S., Guo, T., You, S., Xu, C.: Learning student networks with few data.
  In: AAAI. vol.~34, pp. 4469--4476 (2020)

\bibitem{krizhevsky2012imagenet}
Krizhevsky, A., Sutskever, I., Hinton, G.: Imagenet classification with deep
  convolutional neural networks. NIPS  \textbf{25},  1097--1105 (2012)

\bibitem{lecun2015lenet}
LeCun, Y., et~al.: {LeNet-5: convolutional neural networks}. URL: http://yann.
  lecun. com/exdb/lenet  \textbf{20}(5), ~14 (2015)

\bibitem{lee2019graph}
Lee, S., Song, B.C.: Graph-based knowledge distillation by multi-head attention
  network. arXiv preprint arXiv:1907.02226  (2019)

\bibitem{lopes2017data}
Lopes, R.G., Fenu, S., Starner, T.: Data-free knowledge distillation for deep
  neural networks. arXiv preprint arXiv:1710.07535  (2017)

\bibitem{ma2021undistillable}
Ma, H., Chen, T., Hu, T.K., You, C., Xie, X., Wang, Z.: Undistillable: Making a
  nasty teacher that cannot teach students. In: ICLR (2021)

\bibitem{meng2019conditional}
Meng, Z., Li, J., Zhao, Y., Gong, Y.: Conditional teacher-student learning. In:
  ICASSP. pp. 6445--6449. IEEE (2019)

\bibitem{nayak2021effectiveness}
Nayak, G.K., Mopuri, K.R., Chakraborty, A.: {Effectiveness of Arbitrary
  Transfer Sets for Data-free Knowledge Distillation}. In: CVPR. pp. 1430--1438
  (2021)

\bibitem{nayak2019zero}
Nayak, K., Mopuri, R., Shaj, V., Radhakrishnan, B., Chakraborty, A.: Zero-shot
  knowledge distillation in deep networks. In: ICML. pp. 4743--4751 (2019)

\bibitem{nguyen2021knowledge}
Nguyen, D., Gupta, S., Nguyen, T., Rana, S., Nguyen, P., Tran, T., Le, K.,
  Ryan, S., Venkatesh, S.: Knowledge distillation with distribution mismatch.
  In: ECML-PKDD. pp. 250--265 (2021)

\bibitem{passalis2020heterogeneous}
Passalis, N., Tzelepi, M., Tefas, A.: Heterogeneous knowledge distillation
  using information flow modeling. In: CVPR. pp. 2339--2348 (2020)

\bibitem{pouyanfar2018survey}
Pouyanfar, S., Sadiq, S., Yan, Y., Tian, H., Tao, Y., Reyes, P., Shyu, M.L.,
  Chen, S.C., Iyengar, S.: A survey on deep learning: Algorithms, techniques,
  and applications. ACM Computing Surveys  \textbf{51}(5),  1--36 (2018)

\bibitem{santiago2017building}
Santiago, F., Singh, P., Sri, L., et~al.: {Building Cognitive Applications with
  IBM Watson Services: Volume 6 Speech to Text and Text to Speech}. IBM
  Redbooks (2017)

\bibitem{sohn2015learning}
Sohn, K., Lee, H., Yan, X.: Learning structured output representation using
  deep conditional generative models. NIPS pp. 3483--3491 (2015)

\bibitem{sreenu2019intelligent}
Sreenu, G., Durai, S.: Intelligent video surveillance: a review through deep
  learning techniques for crowd analysis. Journal of Big Data  \textbf{6}(1),
  1--27 (2019)

\bibitem{taigman2014deepface}
Taigman, Y., Yang, M., Ranzato, M., Wolf, L.: Deepface: Closing the gap to
  human-level performance in face verification. In: CVPR. pp. 1701--1708 (2014)

\bibitem{tian2020contrastive}
Tian, Y., Krishnan, D., Isola, P.: Contrastive representation distillation. In:
  ICLR (2020)

\bibitem{wang2020neural}
Wang, D., Li, Y., Wang, L., Gong, B.: {Neural Networks Are More Productive
  Teachers Than Human Raters: Active Mixup for Data-Efficient Knowledge
  Distillation from a Blackbox Model}. In: CVPR. pp. 1498--1507 (2020)

\bibitem{wang2021data}
Wang, Z.: Data-free knowledge distillation with soft targeted transfer set
  synthesis. In: AAAI. vol.~35, pp. 10245--10253 (2021)

\bibitem{wang2021zero}
Wang, Z.: {Zero-Shot Knowledge Distillation from a Decision-Based Black-Box
  Model}. In: ICML (2021)

\bibitem{yim2017gift}
Yim, J., Joo, D., Bae, J., Kim, J.: A gift from knowledge distillation: Fast
  optimization, network minimization and transfer learning. In: CVPR. pp.
  4133--4141 (2017)

\bibitem{yin2020dreaming}
Yin, H., Molchanov, P., Alvarez, J., Li, Z., Mallya, A., Hoiem, D., Jha, N.,
  Kautz, J.: Dreaming to distill: Data-free knowledge transfer via
  deepinversion. In: CVPR. pp. 8715--8724 (2020)

\bibitem{yuan2020revisiting}
Yuan, L., Tay, F., Li, G., Wang, T., Feng, J.: Revisiting knowledge
  distillation via label smoothing regularization. In: CVPR. pp. 3903--3911
  (2020)

\bibitem{zhang2019deep}
Zhang, S., Yao, L., Sun, A., Tay, Y.: Deep learning based recommender system: A
  survey and new perspectives. ACM Computing Surveys  \textbf{52}(1),  1--38
  (2019)

\end{thebibliography}

\end{document}